\documentclass[10pt,twocolumn,letterpaper,table,dvipsnames]{article}

\usepackage{iccv}              

\definecolor{iccvblue}{rgb}{0.21,0.49,0.74}
\usepackage[pagebackref,breaklinks,colorlinks,allcolors=iccvblue]{hyperref}

\usepackage[ruled,vlined]{algorithm2e}
\SetKwInput{KwInput}{Input}
\SetKwInput{KwOutput}{Output}
\usepackage{pifont}

\definecolor{tblue}{HTML}{F0F8FF} 
\definecolor{alogrblue}{HTML}{5188d1}
\definecolor{alogrgrey}{HTML}{727880}
\definecolor{alogryellow}{HTML}{a07018}
\definecolor{alogrgreen}{HTML}{095f42}

\title{When Trackers Date Fish: A Benchmark and Framework for Underwater Multiple Fish Tracking}

\author{Weiran Li\textsuperscript{1,2} \hspace{0.2cm}
	Yeqiang Liu\textsuperscript{1} \hspace{0.2cm}
	Qiannan Guo\textsuperscript{1} \hspace{0.2cm}
	Yijie Wei\textsuperscript{1} \hspace{0.2cm}
	Hwa Liang Leo\textsuperscript{2} \hspace{0.2cm}
	Zhenbo Li\textsuperscript{1}\thanks{Corresponding author.}
	\vspace{1.5mm}\\
	\textsuperscript{\rm 1}China Agricultural University\hspace{1cm}
	\textsuperscript{\rm 2}National University of Singapore
	\vspace{1mm}\\
	{\tt\small \{vranlee,yeqiangliu,guoqiannan,yjwei,lizb\}@cau.edu.cn}
	\vspace{1mm}\\
	{\tt\small weiranli@u.nus.edu, bielhl@nus.edu.sg}
}

\begin{document}
\maketitle

\begin{abstract}
Multiple object tracking (MOT) technology has made significant progress in terrestrial applications, but underwater tracking scenarios remain underexplored despite their importance to marine ecology and aquaculture. In this paper, we present \textit{Multiple Fish Tracking Dataset 2025} (MFT25), a comprehensive dataset specifically designed for underwater multiple fish tracking, featuring 15 diverse video sequences with 408,578 meticulously annotated bounding boxes across 48,066 frames. Our dataset captures various underwater environments, fish species, and challenging conditions including occlusions, similar appearances, and erratic motion patterns. Additionally, we introduce \textit{Scale-aware and Unscented Tracker} (SU-T), a specialized tracking framework featuring an \textit{Unscented Kalman Filter} (UKF) optimized for non-linear fish swimming patterns and a novel \textit{Fish-Intersection-over-Union} (FishIoU) matching that accounts for the unique morphological characteristics of aquatic species. Extensive experiments demonstrate that our SU-T baseline achieves state-of-the-art performance on MFT25, with 34.1 HOTA and 44.6 IDF1, while revealing fundamental differences between fish tracking and terrestrial object tracking scenarios. The dataset and codes are released at \textit{\href{https://vranlee.github.io/SU-T/}{https://vranlee.github.io/SU-T/}}.
\end{abstract}

\begin{figure}
	\centering
	\includegraphics[width=\linewidth]{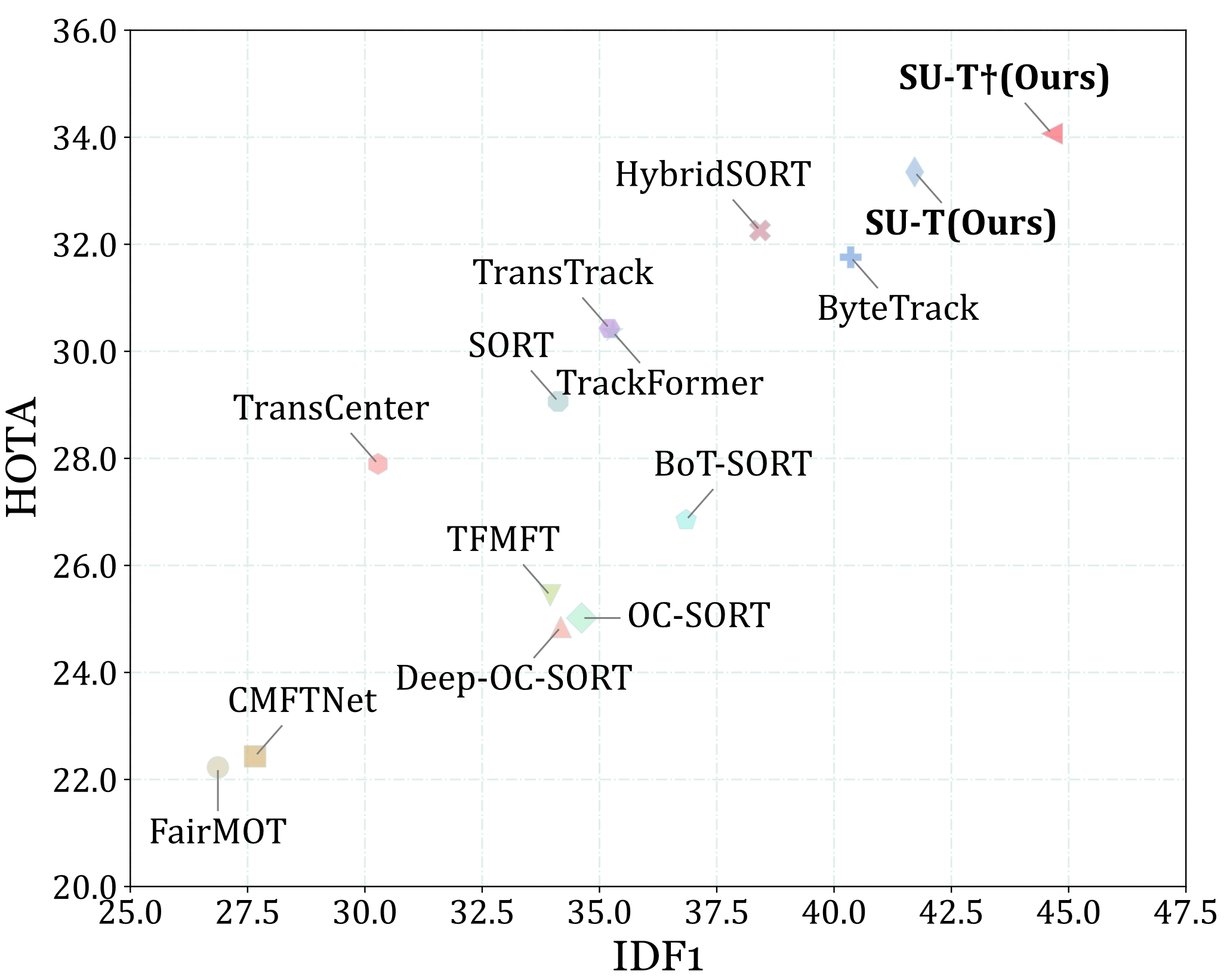}
	\caption{Evaluation of various MFT and MOT methods on MFT25 benchmark. Detailed results are provided in Table~\ref{tab:tracking-comparison}.}
	\label{fig:sota}
\end{figure}

\section{Introduction}
\textit{A tracker's greatest challenge is not merely to find a fish, but to arrange a ${\textbf{date}}$ with the same fleeting shadow—\textbf{a perfect alignment of time and space}.}
\begin{flushright}
	\textit{(Preface)}
\end{flushright}
Fish behavior monitoring and group dynamics analysis form essential technical foundations for marine ecological research, aquaculture optimization, and fishery resource management~\cite{cui2024fish, huang2018fish}. With advancements in computer vision and deep learning, underwater multiple fish tracking (MFT) has emerged as a core technology for efficient, non-invasive observation~\cite{zeng2023fish, li2018real}. It enables quantitative analysis of fish movement patterns, group interactions, and environmental adaptation mechanisms by continuously tracking and associating individual targets across video sequences. MFT offers significant applications in endangered species protection, aquaculture density optimization, and marine ecosystem modeling~\cite{jager2017visual, wang2022real}.

MFT is a specialized application of multiple object tracking (MOT), presents unique challenges in underwater environments~\cite{hassan2024multi, dendorfer2021motchallenge}. It aims to generate continuous trajectories of individual fish through reliable identification across video frames. Unlike single fish tracking, MFT must distinguish between numerous similar-looking fish and maintain consistent identity assignments despite rapid direction changes and frequent occlusions~\cite{bewley2016simple, zhang2022bytetrack}. The ability to resolve confusion between morphologically similar individuals in varying water conditions becomes critical to successful fish tracking, particularly in dense shoaling scenarios where individuals frequently cross paths~\cite{sun2022dancetrack}.

\begin{figure*}[tbp]
	\centering
	\includegraphics[width=\linewidth]{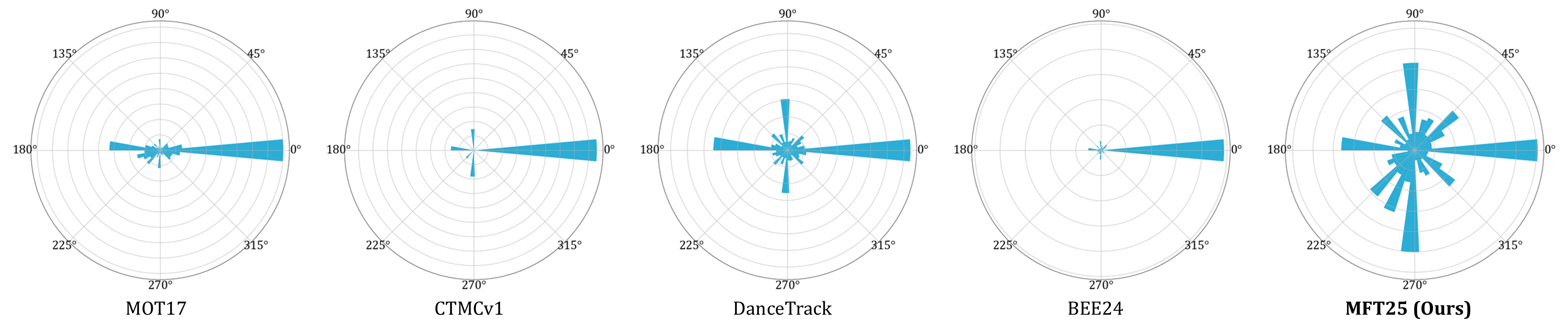}
	\caption{Distribution of target movement directions across datasets. Directional instability is notably more pronounced in the fish dataset compared to other target categories.}
	\label{fig:direct}
\end{figure*}

\begin{figure}
	\centering
	\includegraphics[width=\linewidth]{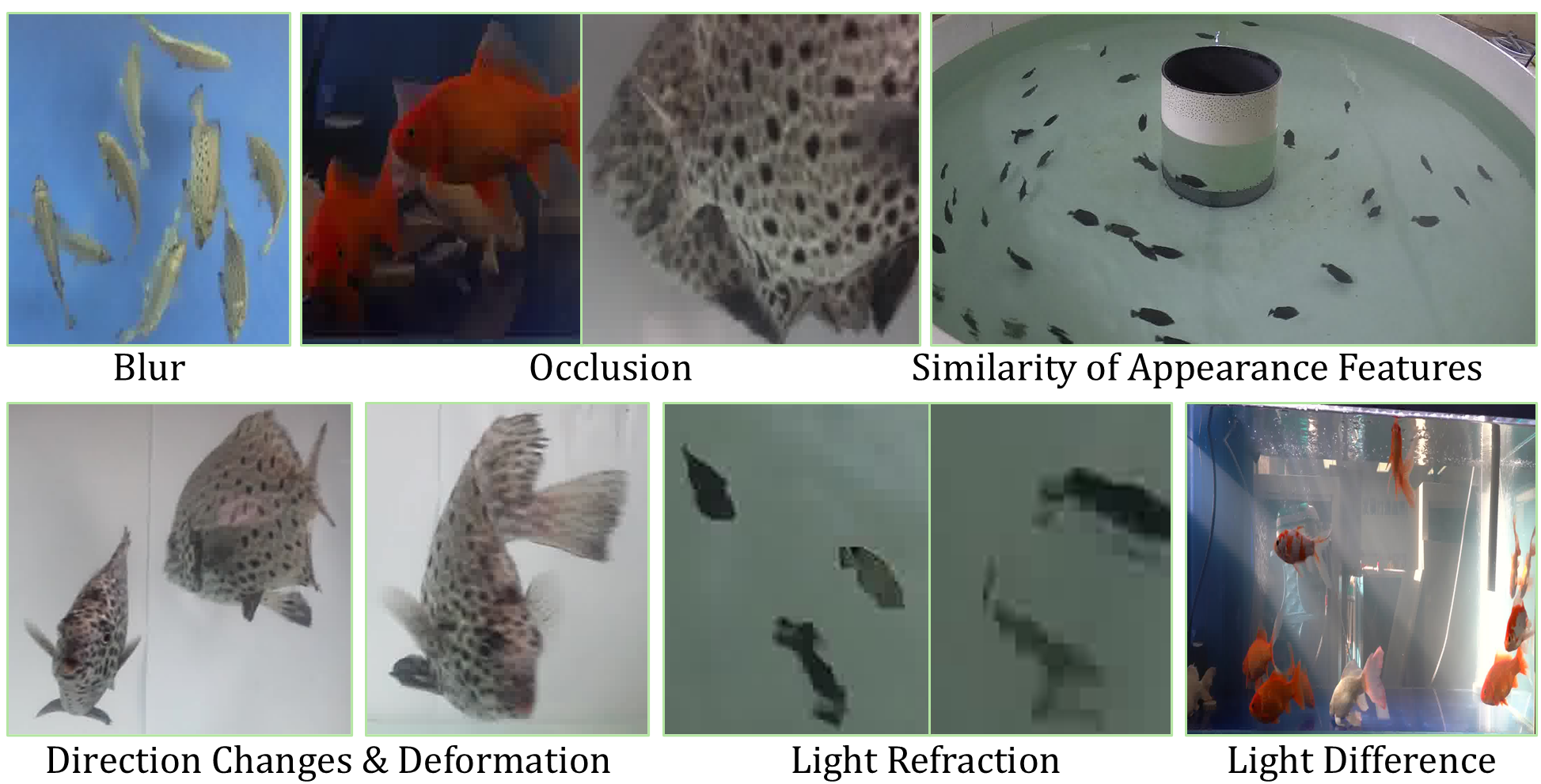}
	\caption{The challenges of multiple fish tracking arise from factors such as fish physiological features and the complexity of underwater scenarios.}
	\label{fig:dataset}
\end{figure}

Current methods largely rely on data-driven approaches, leveraging high-precision detectors to obtain real-time target positions \cite{zhang2021fairmot, li2022cmftnet}. However, tracking fish in complex underwater environments presents several challenges, as shown in Fig.~\ref{fig:dataset}. On the one hand, high morphological similarity among individual fish combined with their erratic movement patterns frequently leads to identity switches and trajectory fragmentation \cite{shevchenko2018fish, li2024tfmft}. On the other hand, existing public datasets suffer from insufficient diversity and poor image quality \cite{pedersen2023brackishmot, pedersen20203d}, limiting the development of tracking models with strong generalization capabilities across complex scenarios.

To address these challenges, we present \textit{Multiple Fish Tracking Dataset 2025} (MFT25), a large-scale dataset specifically designed for underwater MOT task, alongside \textit{Scale-aware and Unscented Tracker} (SU-T), an efficient, lightweight baseline model for online tracking. Our dataset and tracker aim to establish a robust foundation for advancing research in underwater object tracking systems with practical applications in marine ecology and aquaculture. Our main contributions are summarized as follows:
\begin{itemize}
	
	\item We introduce MFT25, a large-scale fish dataset for MOT, featuring 15 diverse video sequences with 408,578 meticulously annotated bounding boxes across 48,066 frames, capturing various underwater environments, fish species, and challenging conditions including occlusions, rapid direction changes, and visually similar appearances.
	
	\item We propose SU-T, a specialized tracking framework featuring an \textit{Unscented Kalman Filter} (UKF) optimized for non-linear fish swimming patterns and a novel \textit{Fish-Intersection-over-Union} (FishIoU) matching that accounts for the unique morphological characteristics and erratic movement behaviors of aquatic species.
	
	\item We conduct extensive comparative experiments demonstrating that our tracker achieves state-of-the-art performance on MFT25, with 34.1 HOTA and 44.6 IDF1, as illustrated in Fig.~\ref{fig:sota}. 
	
	\item Through quantitative analysis, we highlight the fundamental differences between fish tracking and land-based object tracking scenarios, as shown in Fig.~\ref{fig:direct}.
	
\end{itemize}

\section{MFT25: When Trackers Date Fish}
\subsection{Toward Fish Tracking}
\paragraph{Related Methods}
Fish tracking presents unique challenges due to complex underwater environments, distinctive morphology, and erratic swimming behaviors~\cite{cui2024fish}. Early approaches used traditional techniques like background subtraction~\cite{shevchenko2018fish} and object segmentation~\cite{huang2018fish}, while recent advances employ deep learning methods including SiamRPN~\cite{wang2022real}, appearance-based models~\cite{li2018real}, graph-based tracking~\cite{jager2017visual}, and Swin Transformers~\cite{zeng2023fish}. While sharing principles with terrestrial tracking, underwater MOT remains less developed~\cite{hassan2024multi}. Terrestrial MOT research has explored multi-modal fusion~\cite{li2024glatrack}, adaptive frame rates~\cite{liu2023collaborative, feng2023towards}, computational efficiency~\cite{liu2023fasttrack}, appearance modeling~\cite{seidenschwarz2023simple}, depth integration~\cite{liu2025sparsetrack}, and high-density scenarios~\cite{lei2024densetrack}.

Contemporary frameworks fall into three categories: JDE methods use unified networks achieving training efficiency but compromising specialization~\cite{wang2020towards, zhang2021fairmot, li2022cmftnet}; Transformer approaches leverage attention mechanisms with high performance but substantial overhead~\cite{dosovitskiy2020image, carion2020end, xu2022transcenter, meinhardt2022trackformer, li2024tfmft, zhao2024detrs}; SDE methods extend SORT~\cite{bewley2016simple}, decoupling detectors from appearance models for balanced accuracy and efficiency~\cite{zhang2022bytetrack, cao2023observation, xiao2024mambatrack, fischer2023qdtrack, aharon2022bot, yang2024hybrid}. Recent innovations include camera pose estimation~\cite{yi2024ucmctrack}, generative diffusion models~\cite{luo2024diffusiontrack}, and pixel-wise trajectory propagation~\cite{zhao2022tracking}, but show limited underwater adaptability and insufficient real-time performance for practical fish tracking applications.

\begin{figure*}[tb]
	\centering
	\includegraphics[width=\linewidth]{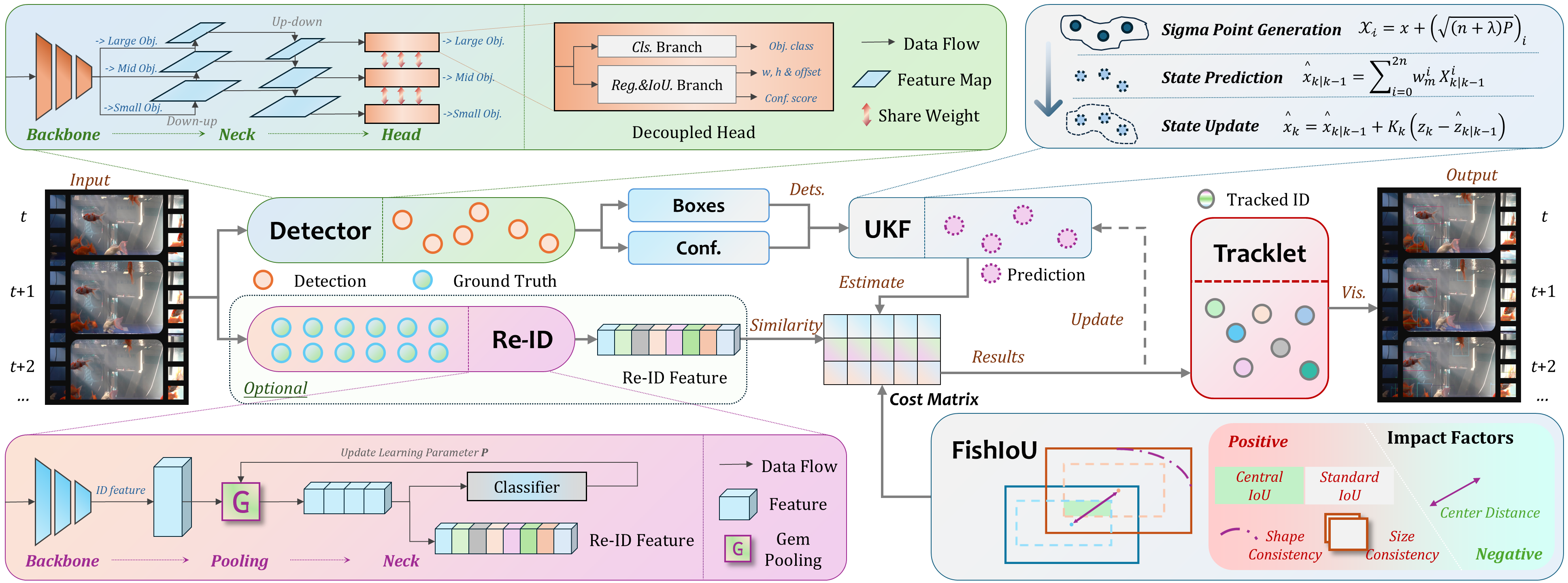}
	\caption{The framework of SU-T. The pipeline consists of three main components: a detector, an association module, and an optional Re-ID module. The detector generates bounding boxes and confidence scores for each frame. The optional Re-ID module extracts feature embeddings to enhance tracking accuracy. The association module uses the FishIoU to calculate matching costs between detected boxes and predicted boxes from the UKF.}
	\label{fig:baseline}
\end{figure*}

\paragraph{Related Datasets}
MOT datasets require frame-by-frame bounding box annotations with consistent identity information across extended video sequences, representing a significant annotation challenge~\cite{dendorfer2020mot20, milan2016mot16, liu2024trafficmot}. Current MOT research predominantly focuses on terrestrial domains, resulting in well-established benchmarks for humans, vehicles, and animals, such as the MOT challenge series~\cite{dendorfer2021motchallenge}, DanceTrack~\cite{sun2022dancetrack}, BEE24~\cite{cao2025topic}, and CTMC~\cite{anjum2020ctmc}.

While several fish-oriented video datasets exist, including Fish4Knowledge, DeepFish, and SeaCLEF~\cite{cui2024fish}, they present significant limitations for modern tracking applications. These early datasets typically suffer from low resolution, poor visibility conditions that obscure fish identities, and inconsistent annotation formats that impede effective model training. Furthermore, other specialized datasets, such as FishTrack23~\cite{dawkins2024fishtrack23} and WebUOT-1M~\cite{zhang2024webuot}, are designed for Single Object Tracking (SOT), a fundamentally different task that provides annotations only for the initialized target. In contrast, MOT datasets require frame-by-frame annotations for all visible targets, along with consistent identity labels to enable long-term association. The CFC dataset~\cite{kay2022caltech} utilizes sonar imaging, which represents a distinct data modality from the optical videos commonly used in aquaculture applications. More recent standardized fish tracking datasets, including BrackishMOT~\cite{pedersen2023brackishmot}, 3D-ZeF~\cite{pedersen20203d}, and MFT22~\cite{li2024tfmft}, have emerged with consistent annotation protocols. However, these datasets remain limited in both environmental diversity and scale, typically featuring simplified scenarios under controlled conditions. The absence of comprehensive, high-quality, and standardized fish tracking datasets thus represents a critical bottleneck that constrains significant advances in underwater MFT research.

\subsection{Dataset Construction}
The MFT25 dataset was captured using imaging equipment of Canon $EOSR6$ and Sony $\alpha7M3$, across diverse aquaculture environments. Recording locations encompassed both industrial circulating water aquaculture ponds and controlled laboratory tanks to ensure environmental diversity. The dataset features multiple fish species with distinctly different morphologies, including commercially valuable groupers and ornamental koi at various developmental stages, providing substantial variation in appearance characteristics.

To ensure comprehensive scenario coverage, we systematically deployed multiple camera configurations, including both overhead and horizontal perspectives, across varied illumination conditions from daylight to nocturnal settings. The dataset consists exclusively of authentic footage without synthetic augmentation, preserving the natural complexity of scenarios. All bounding box annotations were created using DarkLabel\footnote{\url{https://github.com/darkpgmr/DarkLabel}} software through manual selection and verification processes. The resulting MFT25 dataset encompasses 15 diverse video sequences containing 223 distinct fish trajectories across 48,066 frames, with a total of 408,578 precisely annotated bounding boxes. This represents a substantial advancement in scale, containing 2.6-8.3 times more annotated instances compared to previous fish tracking datasets. Table~\ref{tab:dataset-comparison} presents a comprehensive statistical comparison with existing fish tracking benchmarks.

\begin{table}[tbp]
	\caption{Quantitative comparison of MFT datasets.}
	\label{tab:dataset-comparison}
	\centering
	\resizebox{\linewidth}{!}{%
		\begin{tabular}{l|cccc}
			\toprule
			\textbf{Dataset} & \textbf{BrackishMOT} & \textbf{3D-ZeF} & \textbf{MFT22} & \textbf{MFT25(Ours)} \\
			\midrule
			Clips & 98 & 8 & 10 & 15 \\
			Tracks & 638 & 32 & 234 & 223 \\
			FPS & 25 & 60 & 25 & 25 \\
			\midrule
			\rowcolor{tblue} \textbf{Frames} & 14,017 & 14,398 & 9,100 & \textbf{48,066} \\
			\rowcolor{tblue} \textbf{Boxes} & 49,364 & 86,452 & 155,437 & \textbf{408,578} \\
			\bottomrule
		\end{tabular}%
	}
\end{table}

\section{SU-T: A MFT Baseline}
\subsection{Framework}

To address the unique challenges of MFT, we propose \textit{Scale-aware and Unscented Tracker} (SU-T), a specialized baseline following the SDE paradigm. As illustrated in Fig.~\ref{fig:baseline}, our framework comprises three primary components: a detector, an association module, and an optional Re-Identification (Re-ID) module. The processing pipeline begins with video frames being fed into the detector, which generates bounding boxes with corresponding confidence scores. These detections are then processed by the association module, where our specialized FishIoU metric calculates matching costs between detected boxes and predictions from the UKF. The Hungarian algorithm performs optimal assignment to update existing trajectories and establish new tracks when necessary. To address the challenge of visually similar fish, SU-T integrates an optional Re-ID module that extracts discriminative feature embeddings. These embeddings work synergistically with the FishIoU metric during association, significantly enhancing tracking accuracy by maintaining consistent identities even when fish exhibit nearly identical appearances.

\subsection{Detector and Re-ID}
Considering the variability of fish movements and the significant scale variations due to varying distances from the camera, our tracker adopts a mainstream pyramid-based design following~\cite{zhang2022bytetrack, cao2023observation, yang2024hybrid} and employs decoupled heads to predict point centers, bounding boxes, offsets, and confidence scores~\cite{ge2021yolox}. The different decoupled heads share feature parameters across pyramid levels. Additionally, an optional re-identification module continuously updates GeM Pooling through learnable parameters, outputting target appearance features that are incorporated into the cost matrix for subsequent association, thereby providing robust appearance similarity cues for tracking.

\subsection{Unscented Kalman Filter}

\begin{figure}[tbp]
	\centering
	\includegraphics[width=0.95\linewidth]{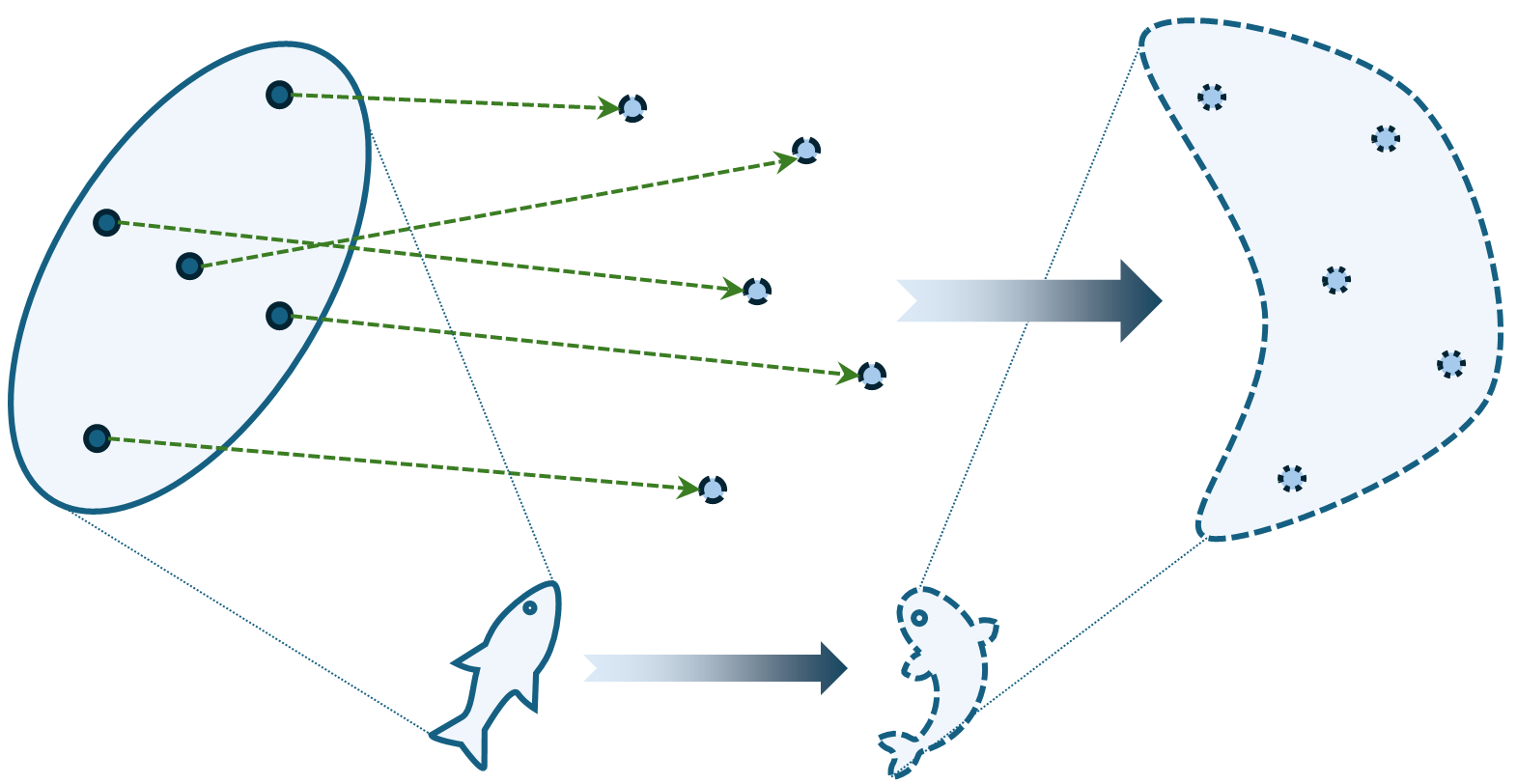}
	\caption{\textit{Unscented Kalman Filter} (UKF) motion model used in SU-T for predicting fish movement in complex underwater environments.}
	\label{fig:ukf}
\end{figure}

The \textit{Unscented Kalman Filter} (UKF) is particularly well-suited for tracking fish due to their non-linear motion patterns, as shown in Fig.~\ref{fig:ukf}. Unlike standard Kalman Filter, UKF uses a deterministic sampling technique to handle non-linearities. The three core mathematical components of our UKF implementation are as follows.

\subsubsection{Sigma Points Generation}
For state vector $\mathbf{x} \in \mathbb{R}^n$ with covariance $\mathbf{P}$, we generate $2n+1$ sigma points:

\begin{equation}
	\mathcal{X}_0 = \mathbf{x}
\end{equation}

\begin{equation}
	\mathcal{X}_i = \mathbf{x} + \left(\sqrt{(n+\lambda)\mathbf{P}}\right)_i, \quad i=1,\ldots,n
\end{equation}

\begin{equation}
	\mathcal{X}_{i+n} = \mathbf{x} - \left(\sqrt{(n+\lambda)\mathbf{P}}\right)_i, \quad i=1,\ldots,n
\end{equation}

\noindent where $\lambda = \alpha^2(n+\kappa)-n$ is a scaling parameter, $\alpha$ controls spread of points, $\kappa$ is a secondary parameter (typically 3-n), and $\left(\sqrt{(n+\lambda)\mathbf{P}}\right)_i$ is the $i$-th column of the matrix square root.

\subsubsection{Prediction Step}
Each sigma point is propagated through the non-linear state transition function $\mathbf{f}$:

\begin{equation}
	\mathcal{X}_{k|k-1}^i = \mathbf{f}(\mathcal{X}_{k-1}^i), \quad i=0,\ldots,2n
\end{equation}

\begin{equation}
	\hat{\mathbf{x}}_{k|k-1} = \sum_{i=0}^{2n} w_m^i \mathcal{X}_{k|k-1}^i
\end{equation}

\begin{equation}
	\mathbf{P}_{k|k-1} = \sum_{i=0}^{2n} w_c^i [\mathcal{X}_{k|k-1}^i - \hat{\mathbf{x}}_{k|k-1}][\mathcal{X}_{k|k-1}^i - \hat{\mathbf{x}}_{k|k-1}]^T + \mathbf{Q}_k
\end{equation}

\noindent where $w_m^i$ and $w_c^i$ are weight coefficients for mean and covariance, and $\mathbf{Q}_k$ is the process noise covariance.

\subsubsection{Measurement Update Step}
The predicted sigma points are transformed through the measurement function $\mathbf{h}$:

\begin{equation}
	\mathcal{Z}_{k|k-1}^i = \mathbf{h}(\mathcal{X}_{k|k-1}^i), \quad i=0,\ldots,2n
\end{equation}

\begin{equation}
	\hat{\mathbf{z}}_{k|k-1} = \sum_{i=0}^{2n} w_m^i \mathcal{Z}_{k|k-1}^i
\end{equation}

\begin{equation}
	\mathbf{P}_{zz} = \sum_{i=0}^{2n} w_c^i [\mathcal{Z}_{k|k-1}^i - \hat{\mathbf{z}}_{k|k-1}][\mathcal{Z}_{k|k-1}^i - \hat{\mathbf{z}}_{k|k-1}]^T + \mathbf{R}_k
\end{equation}

\begin{equation}
	\mathbf{P}_{xz} = \sum_{i=0}^{2n} w_c^i [\mathcal{X}_{k|k-1}^i - \hat{\mathbf{x}}_{k|k-1}][\mathcal{Z}_{k|k-1}^i - \hat{\mathbf{z}}_{k|k-1}]^T
\end{equation}

\begin{equation}
	\mathbf{K}_k = \mathbf{P}_{xz}\mathbf{P}_{zz}^{-1}
\end{equation}

\begin{equation}
	\hat{\mathbf{x}}_k = \hat{\mathbf{x}}_{k|k-1} + \mathbf{K}_k(\mathbf{z}_k - \hat{\mathbf{z}}_{k|k-1})
\end{equation}

\begin{equation}
	\mathbf{P}_k = \mathbf{P}_{k|k-1} - \mathbf{K}_k\mathbf{P}_{zz}\mathbf{K}_k^T
\end{equation}

\noindent where $\mathbf{z}_k$ is the actual measurement, $\mathbf{R}_k$ is the measurement noise covariance, and $\mathbf{K}_k$ is the Kalman gain.

\subsection{FishIoU: Scale-aware Association}

The unique morphology and movement patterns of fish present significant challenges for standard object association. We introduce \textit{Fish-Intersection-over-Union} (FishIoU), a specialized association IoU that accounts for the elongated body structure, erratic motion patterns, and size variations common in fish species.

Given two bounding boxes $B_1 = [x_1, y_1, x_2, y_2]$ and $B_2 = [x'_1, y'_1, x'_2, y'_2]$, we first compute the standard IoU as:

\begin{equation}
	\text{IoU} = \frac{|B_1 \cap B_2|}{|B_1 \cup B_2|}
\end{equation}

\noindent where $|B_1 \cap B_2|$ represents the intersection area and $|B_1 \cup B_2|$ the union area. Then, to account for fish morphology, we incorporate a center distance penalty:

\begin{equation}
	d_c = \frac{(c_x - c'_x)^2 + (c_y - c'_y)^2}{d_{\text{diag}}^2}
\end{equation}

\noindent where $(c_x, c_y)$ and $(c'_x, c'_y)$ are the centers of $B_1$ and $B_2$ respectively, and $d_{\text{diag}}^2$ is the squared diagonal length of the enclosing box. Considering fish bodies typically have an elongated structure with important features concentrated in the front, we define a central region for each box with asymmetric insets to emphasize this characteristic:

\begin{equation}
	B_1^c = [x_1 + \alpha w_1,\, y_1 + \beta h_1,\, x_2 - \gamma w_1,\, y_2 - \beta h_1]
\end{equation}
\begin{equation}
	B_2^c = [x'_1 + \alpha w_2,\, y'_1 + \beta h_2,\, x'_2 - \gamma w_2,\, y'_2 - \beta h_2]
\end{equation}

\noindent where $w_1, h_1$ and $w_2, h_2$ are the width and height of $B_1$ and $B_2$ respectively, and $\alpha, \beta, \gamma$ are constant factors determined empirically based on fish morphological characteristics. By default, $\alpha = 0.15$, $\beta = 0.3$, and $\gamma = 0.25$. The central IoU is then calculated as:

\begin{equation}
	\text{cIoU} = \frac{|B_1^c \cap B_2^c|}{|B_1^c \cup B_2^c|}
\end{equation}

To account for consistent fish orientation, we consider the aspect ratio consistency:

\begin{equation}
	\alpha_r = \frac{\min(r_1, r_2)}{\max(r_1, r_2)}
\end{equation}

\noindent where $r_1 = \frac{w_1}{h_1}$ and $r_2 = \frac{w_2}{h_2}$ are the aspect ratios of the two boxes.

Additionally, we incorporate area ratio consistency, since the size of the fish does not change abruptly between frames:

\begin{equation}
	\alpha_a = \frac{\min(a_1, a_2)}{\max(a_1, a_2)}
\end{equation}

\noindent where $a_1 = w_1 \times h_1$ and $a_2 = w_2 \times h_2$ are the areas of the two boxes.

For small targets, we apply a scale factor to reduce the center distance penalty:

\begin{equation}
	s = 1 - e^{-\frac{\min(a_1, a_2)}{1000}}
\end{equation}

The final FishIoU metric combines these components with specific weights optimized for fish tracking:

\begin{equation}
	\text{FishIoU} = \omega_1 \cdot \text{IoU} + \omega_2 \cdot \text{cIoU} + \omega_3 \cdot \alpha_r + \omega_4 \cdot \alpha_a - \omega_5 \cdot s \cdot d_c
\end{equation}

\noindent where $\omega_1 = 1.0$, $\omega_2 = 0.3$, $\omega_3 = 0.1$, $\omega_4 = 0.2$, and $\omega_5 = 0.4$ are weights determined empirically through extensive experiments.

\subsection{Assocation}

Our association strategy employs progressive confidence-based processing that significantly reduces identity switches while maintaining computational efficiency. Algorithm~\ref{alg:tracking} presents the complete multi-level cascade tracking process, which integrates our UKF motion prediction and FishIoU matching with a cascaded association strategy for aquatic environments.

Building upon frameworks from HybridSORT~\cite{yang2024hybrid}, we adopt dual-confidence matching of ByteTrack~\cite{zhang2022bytetrack} and heuristic observation-centric recovery of OC-SORT~\cite{cao2023observation}, and implements three association stages. In the first stage, high-confidence detections are matched with existing tracks using our specialized FishIoU, establishing reliable primary associations even when fish exhibit rapid direction changes. The second stage associates remaining tracks with low-confidence detections, effectively recovering temporarily occluded targets while filtering false positives induced by water turbidity and reflections. The final stage attempts to reconnect tracks with their historical appearances, addressing the frequent, abrupt directional changes and non-linear swimming patterns characteristic of fish locomotion.

For cost calculation, our framework integrates both spatial and appearance information when available. Spatially, the FishIoU outperforms standard IoU by incorporating fish-specific morphological features into the matching process. When enabled, the Re-ID module provides discriminative appearance embeddings that effectively differentiate between visually similar individuals swimming in close proximity.

\begin{algorithm}[!t]
	\caption{Multi-Level Cascade Tracking}
	\label{alg:tracking}
	\SetAlgoLined
	
	\KwInput{Detections $\mathcal{D}$ with scores, Existing tracks $\mathcal{T}$}
	\KwOutput{Updated tracks $\mathcal{T}$}
	
	\textcolor{alogryellow}{/* Prediction */}\\
	\For{$T_j \in \mathcal{T}$}{
		\textcolor{alogrblue}{$\hat{B}_j, s_j \leftarrow$ UKF.predict($T_j$)} \textcolor{alogrgrey}{/* Predict box and score */}\\
	}
	
	\textcolor{alogryellow}{/* First Association: High-confidence */}\\
	$\mathcal{D}_{high} \leftarrow \{d_i \in \mathcal{D} : score(d_i) > \tau_{high}\}$\\
	\textcolor{alogrblue}{$\mathbf{C} \leftarrow$ FishIoU($\mathcal{D}_{high}$, $\{\hat{B}_j\}$)} \textcolor{alogrgrey}{/* Base cost */}\\
	\textcolor{alogrgreen}{\If{Use Re-ID}{$\mathbf{C} \leftarrow \omega_1 \mathbf{C} + \omega_2$ EmbeddingDistance($\mathcal{D}_{high}$, $\mathcal{T}$)}}
	$\mathcal{M}_1, \mathcal{U}_{\mathcal{D}}, \mathcal{U}_{\mathcal{T}} \leftarrow$ Hungarian($-\mathbf{C}$)\\
	
	\For{$(i,j) \in \mathcal{M}_1$}{$T_j$.update($d_i$)}
	
	\textcolor{alogryellow}{/* Second Association: Low-confidence */}\\
	$\mathcal{D}_{low} \leftarrow \{d_i \in \mathcal{D} : \tau_{low} < score(d_i) < \tau_{high}\}$\\
	\textcolor{alogrblue}{$\mathbf{C}_{iou} \leftarrow$ FishIoU($\mathcal{D}_{low}$, $\{\hat{B}_j : T_j \in \mathcal{U}_{\mathcal{T}}\}$)}\\
	\textcolor{alogrgreen}{\If{Use Re-ID}{$\mathbf{C}_{iou} \leftarrow \mathbf{C}_{iou} + \lambda \cdot$ EmbeddingDistance($\mathcal{D}_{low}$, $\{T_j \in \mathcal{U}_{\mathcal{T}}\}$)}}
	$\mathcal{M}_2 \leftarrow$ Hungarian($-\mathbf{C}_{iou}$)\\
	
	\For{$(i,j) \in \mathcal{M}_2$ where $\mathbf{C}_{iou}[i,j] > \tau_{iou}$}{
		$T_j$.update($\mathcal{D}_{low}[i]$) \textcolor{alogrgrey}{/* Update state */}\\
		$\mathcal{U}_{\mathcal{T}} \leftarrow \mathcal{U}_{\mathcal{T}} \setminus \{T_j\}$\\
	}
	
	\textcolor{alogryellow}{/* Third Association: Last-chance */}\\
	\textcolor{alogrblue}{$\mathbf{C}_{last} \leftarrow$ FishIoU($\mathcal{D}_{high}[\mathcal{U}_{\mathcal{D}}]$, $\{T_j.last\_observation : T_j \in \mathcal{U}_{\mathcal{T}}\}$)}\\
	\For{$(i,j) \in$ Hungarian($-\mathbf{C}_{last}$) where $\mathbf{C}_{last}[i,j] > \tau_{iou}$}{
		$T_j$.update($\mathcal{D}_{high}[\mathcal{U}_{\mathcal{D}}[i]]$)\\
		Remove $i$ from $\mathcal{U}_{\mathcal{D}}$, $T_j$ from $\mathcal{U}_{\mathcal{T}}$\\
	}
	
	\textcolor{alogryellow}{/* Finalize */}\\
	Update all $T_j \in \mathcal{U}_{\mathcal{T}}$ without observation\\
	Initialize new tracks from $\mathcal{D}_{high}[\mathcal{U}_{\mathcal{D}}]$\\
	Remove tracks with $time\_since\_update > max\_age$\\
	
	\Return $\mathcal{T}$
\end{algorithm}

\section{Experiments}
\subsection{Implementation Details}

For all experiments, we employed the same YOLOX~\cite{ge2021yolox} detector and the same Re-ID network~\cite{he2023fastreid} for SDE-based models, trained with consistent hyperparameter configurations following the established protocols from ByteTrack~\cite{zhang2022bytetrack} and BoT-SORT~\cite{aharon2022bot}. All models were trained on the MFT25 training set using a NVIDIA A100 GPU, with performance evaluated on the test set using standard MOT metrics, including the comprehensive HOTA~\cite{luiten2021hota} metric alongside traditional CLEAR~\cite{bernardin2008evaluating} metrics such as MOTA, IDF1, and ID switches (IDs).

\subsection{Benchmark Results}

\begin{table*}[tbp]
	\caption{Comparison of different tracking methods on the MFT25 dataset. $^\dagger$ indicates the integration of Re-ID module. The best two results are bolded and underlined respectively. Same as follows.}
	\label{tab:tracking-comparison}
	\centering
	\resizebox{\linewidth}{!}{%
		\begin{tabular}{l|c|c|cccccccc}
			\toprule
			\textbf{Method}                                 & \textbf{Class} &   \textbf{HOTA↑}   &   \textbf{IDF1↑}   &   \textbf{MOTA↑}   &   \textbf{AssA↑}   &   \textbf{DetA↑}   &  \textbf{IDs↓}  &  \textbf{IDFP↓}   &  \textbf{IDFN↓}   &  \textbf{Frag↓}  \\ \midrule
			FairMOT~\cite{zhang2021fairmot}                 &      JDE       &       22.226       &       26.867       &       47.509       &       13.910       &       35.606       &       939       &       58198       &      113393       &       3768       \\
			CMFTNet~\cite{li2022cmftnet}                    &      JDE       &       22.432       &       27.659       &       46.365       &       14.278       &       35.452       &      1301       &       64754       &      111263       &       2769       \\
			Deep-OC-SORT~\cite{maggiolino2023deep}          &      SDE       &       24.848       &       34.176       &       46.721       &       17.537       &       35.373       &       550       & \underline{53478} &      104024       &       3659       \\
			OC-SORT~\cite{cao2023observation}               &      SDE       &       25.017       &       34.620       &       46.706       &       17.783       &       35.369       &       550       &  \textbf{52934}   &      103495       &       3651       \\
			TFMFT~\cite{li2024tfmft}                        &  Transformer   &       25.440       &       33.950       &       49.725       &       17.112       &       38.059       &       719       &       63125       &      102378       &       3251       \\
			BoT-SORT~\cite{aharon2022bot}                   &      SDE       &       26.848       &       36.847       &       49.108       &       19.446       &       37.241       & \underline{500} &       57581       &       99181       &       2704       \\
			TransCenter~\cite{xu2022transcenter}            &  Transformer   &       27.896       &       30.278       &       68.693       &  \textbf{30.255}   &       30.301       &       807       &      101223       &      101002       &       1992       \\
			SORT~\cite{bewley2016simple}                    &      SDE       &       29.063       &       34.119       &       69.038       &       16.952       &       50.195       &       778       &       88928       &       96815       & \underline{1726} \\
			TrackFormer~\cite{meinhardt2022trackformer}     &  Transformer   &       30.361       &       35.285       &  \textbf{74.609}   &       17.661       &  \textbf{52.649}   &       718       &       89391       &       94720       &       1729       \\
			TransTrack~\cite{sun2020transtrack}             &  Transformer   &       30.426       &       35.215       &       68.983       &       18.525       & \underline{50.458} &      1116       &       96045       &       93418       &       2588       \\
			ByteTrack~\cite{zhang2022bytetrack}             &      SDE       &       31.758       &       40.355       & \underline{69.586} &       20.392       &       49.712       &  \textbf{489}   &       80765       &       87866       &  \textbf{1555}   \\
			HybridSORT~\cite{yang2024hybrid}                &      SDE       &       32.258       &       38.421       &       68.905       &       20.936       &       49.992       &       613       &       85924       &       90022       &       1931       \\
			HybridSORT$^\dagger$~\cite{yang2024hybrid}      &      SDE       &       32.705       & \underline{41.727} &       69.167       &       21.701       &       49.697       &       562       &       79189       &       85830       &       1963       \\
			\rowcolor{tblue} \textbf{SU-T (Ours)}           &      SDE       & \underline{33.351} &       41.717       &       68.450       &       22.425       &       49.943       &       607       &       83111       & \underline{84814} &       2006       \\
			\rowcolor{tblue} \textbf{SU-T$^\dagger$ (Ours)} &      SDE       &  \textbf{34.067}   &  \textbf{44.643}   &       68.958       & \underline{23.594} &       49.531       &       544       &       76440       &  \textbf{81304}   &       2011       \\ \bottomrule
		\end{tabular}%
	}
\end{table*}

We compare our proposed SU-T baseline with state-of-the-art MOT and MFT methods on the MFT25 dataset. Table~\ref{tab:tracking-comparison} presents the comprehensive comparison results. Our method achieves the best overall performance with a HOTA score of 34.1, surpassing the previous best method by 1.4. This improvement demonstrates the effectiveness of our specialized fish tracking approach compared to general-purpose tracking methods. The superiority of our method is particularly evident in association metrics, where SU-T achieves the highest IDF1 score of 44.6 with Re-ID module, significantly outperforming other methods. This indicates that our approach is more effective at maintaining consistent fish identities across frames, which is crucial for accurate trajectory analysis in complex underwater scenarios.

\begin{figure}[tbp]
	\centering
	\includegraphics[width=\linewidth]{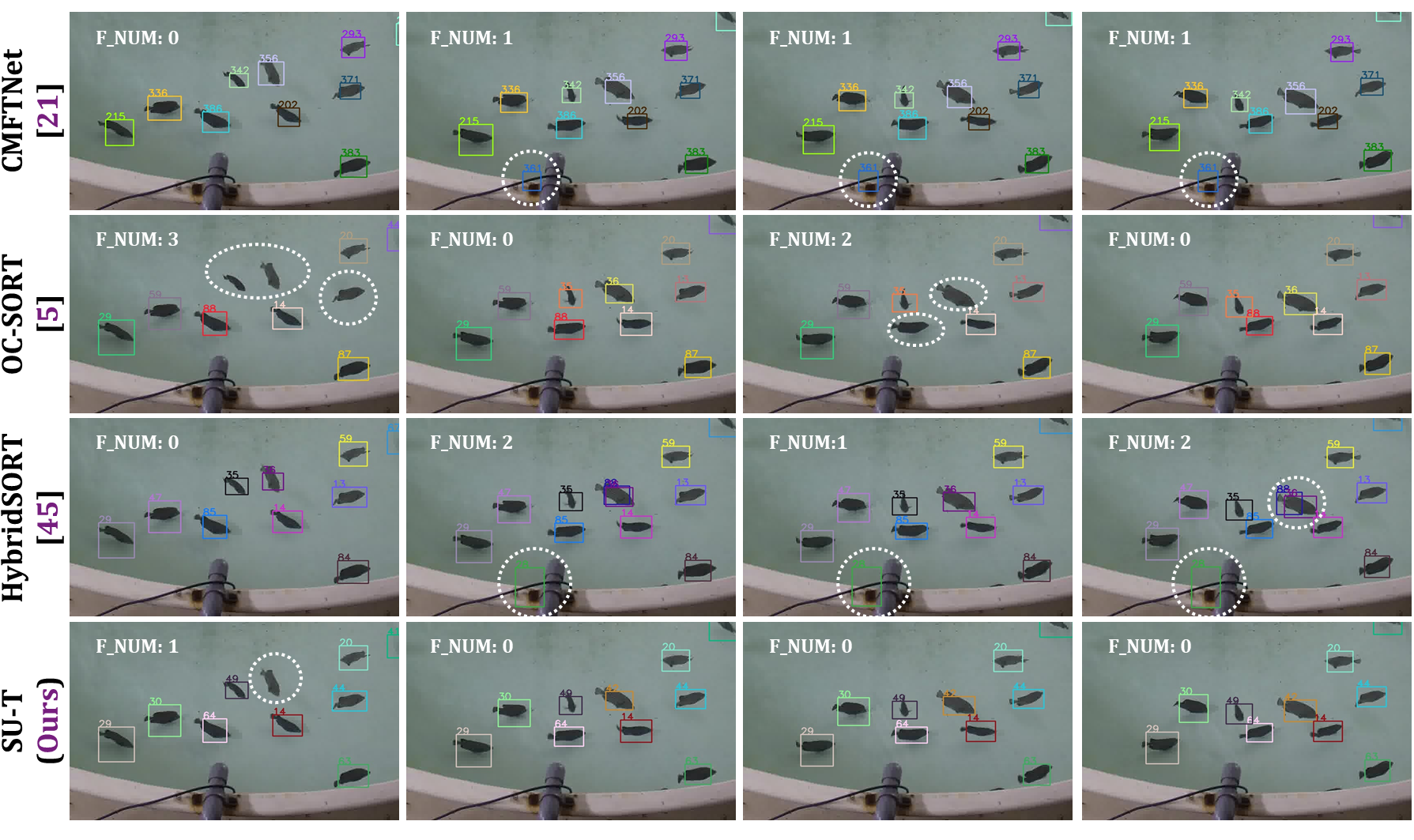}
	\caption{Tracking performance of various trackers on the MFT25 dataset. F\_NUM denotes false tracked number, including IDFN, IDFP, and IDs. Best viewed in color.}
	\label{fig:vis_com}
\end{figure}

Although Transformer-based TrackFormer~\cite{meinhardt2022trackformer} achieving the highest MOTA score of 74.6. However, it exhibit relatively weaker performance in identity preservation metrics such as IDF1 and AssA. Besides, transformer-based trackers impose substantial computational overhead, rendering them impractical for real-time underwater monitoring applications. The performance gap between conventional terrestrial-focused trackers and SU-T validates our hypothesis that underwater tracking scenarios necessitate domain-specific adaptations. Visual comparisons of various tracking methods on MFT25 are illustrated in Fig.~\ref{fig:vis_com}. 

In addition, we conducted additional experiments to evaluate the generalization capability of SU-T on mainstream land-based tracking benchmarks MOT17 and MOT20~\cite{dendorfer2021motchallenge}. Our baseline achieved 60.4 and 56.5 HOTA, respectively.\footnote{The results have been registered at \url{https://motchallenge.net/}}

\subsection{Ablation Studies}

\begin{table}[tbp]
	\caption{Results on different Re-ID models with standard Kalman Filter and IoU association.}
	\label{tab:backbone-ablation}
	\centering
	\resizebox{\linewidth}{!}{%
		\begin{tabular}{l|c|c|cccc}
			\toprule
			\textbf{Method} & \textbf{IBN} & \textbf{HOTA↑} & \textbf{IDF1↑} & \textbf{MOTA↑} & \textbf{AssA↑} & \textbf{IDs↓} \\
			\midrule
			SBS-R50 &  & 30.950 & 39.937 & 68.780 & 19.599 & 713 \\
			SBS-R50 & \ding{51} & 30.560 & 37.919 & 68.909 & 19.010 & 659 \\
			SBS-R101 &  & 30.270 & 38.104 & 68.796 & 18.599 & 678 \\
			SBS-R101 & \ding{51} & 30.684 & 39.996 & 68.912 & 19.134 & 638 \\
			SBS-S50 &  & \underline{32.705} & \underline{41.727} & \underline{69.167} & \underline{21.701} & 562 \\
			SBS-S50 & \ding{51} & 32.412 & 40.977 & 69.030 & 21.183 & \underline{558} \\
			SBS-S101 &  & \textbf{33.842} & \textbf{43.748} & 69.043 & \textbf{23.154} & \textbf{550} \\
			SBS-S101 & \ding{51} & 31.900 & 40.201 & \textbf{69.212} & 20.610 & 584 \\
			\bottomrule
		\end{tabular}%
	}
\end{table}

We conducted extensive ablation experiments to evaluate the components. Table~\ref{tab:backbone-ablation} shows the impact of different Re-ID model on tracking performance. The SBS-S101~\cite{he2023fastreid} model achieves the best overall performance with 33.8 HOTA. Notably, the Instance-Batch Normalization (IBN)~\cite{pan2018two} variants do not consistently improve performance, suggesting that domain adaptation techniques optimized for terrestrial tracking may not transfer effectively to underwater scenarios. As a result, we adopt SBS-S101 as our Re-ID module.

Table~\ref{tab:iou-ablation} compares different IoU for association cost calculation. Our specialized FishIoU outperforms other association methods, achieving the highest 33.4 HOTA and 41.7 IDF1 scores. When combined with Re-ID module, performance further improves across all metrics, confirming that our FishIoU better handles the unique morphology and movement patterns of fish species.

\begin{table}[tbp]
	\caption{Ablation study comparing different IoU for association cost calculation. Center and IoU represent the center points distance and the standard IoU, respectively.}
	\label{tab:iou-ablation}
	\centering
	\resizebox{\linewidth}{!}{%
		\begin{tabular}{l|c|ccc|ccc}
			\toprule
			\textbf{Method} & \textbf{HOTA↑} & \textbf{IDF1↑} & \textbf{MOTA↑} & \textbf{AssA↑} & \textbf{IDs↓} & \textbf{IDFP↓} & \textbf{IDFN↓} \\
			\midrule
			Center & 28.865 & 37.348 & 66.313 & 17.410 & 1273 & 88585 & 91297 \\
			IoU & 32.790 & 40.098 & 68.839 & 21.573 & 579 & 84648 & 87347 \\
			CIoU~\cite{zheng2021enhancing} & 30.720 & 39.598 & 67.425 & 19.325 & 727 & 87422 & 87556 \\
			DIoU~\cite{zheng2020distance} & 30.764 & 39.575 & 67.519 & 19.326 & 728 & 87344 & 87617 \\
			HMIoU~\cite{yang2024hybrid} & 32.258 & 38.421 & \underline{68.905} & 20.936 & 613 & 85924 & 90022 \\
			GIoU~\cite{rezatofighi2019generalized} & 32.885 & 39.957 & 68.798 & 21.686 & \underline{573} & 84896 & 87538 \\
			\rowcolor{tblue} \textbf{FishIoU} & \underline{33.351} & \underline{41.717} & 68.450 & \underline{22.425} & 607 & \underline{83111} & \underline{84814} \\
			\rowcolor{tblue} \textbf{FishIoU$^\dagger$} & \textbf{33.581} & \textbf{43.268} & \textbf{68.989} & \textbf{22.779} & \textbf{547} & \textbf{78473} & \textbf{83243} \\
			\bottomrule
		\end{tabular}%
	}
\end{table}

Table~\ref{tab:motion-ablation} investigates the effectiveness of different motion models combined with our association metrics. The UKF consistently outperforms other motion models including standard Kalman Filter (KF), Adaptive Kalman Filter (AKF), and Strong Tracking Filter (STF) across both HMIoU~\cite{yang2024hybrid} and FishIoU association. This validates our hypothesis that non-linear motion models are more appropriate for capturing the complex swimming patterns of fish. The best performance is achieved by combining UKF with FishIoU and Re-ID, resulting in a HOTA of 34.1 and IDF1 of 44.6.

\begin{table}[tbp]
	\caption{Ablation study on different motion models and association IoUs.}
	\label{tab:motion-ablation}
	\centering
	\resizebox{\linewidth}{!}{%
		\begin{tabular}{l|cc|c|cccc}
			\toprule
			\textbf{Method} & \textbf{HMIoU} & \textbf{FishIoU} & \textbf{HOTA↑} & \textbf{IDF1↑} & \textbf{MOTA↑} & \textbf{AssA↑} & \textbf{IDs↓} \\
			\midrule
			KF & \ding{51} & & 32.258 & \underline{38.421} & 68.905 & 20.936 & 613 \\
			AKF & \ding{51} & & 28.769 & 33.689 & 67.827 & 16.954 & 1031 \\
			STF & \ding{51} & & 31.105 & 36.911 & \textbf{69.161} & 19.445 & 667 \\
			\rowcolor{tblue} \textbf{UKF} & \ding{51} & & \underline{32.406} & 38.408 & 68.933 & \underline{21.096} & \underline{609} \\
			\rowcolor{tblue} \textbf{UKF$^\dagger$} & \ding{51} & & \textbf{33.737} & \textbf{43.880} & \underline{69.057} & \textbf{23.063} & \textbf{528} \\
			\midrule
			KF & & \ding{51} & 33.051 & 41.041 & 68.503 & 22.022 & 612 \\
			AKF & & \ding{51} & 22.551 & 24.017 & 65.535 & 10.682 & 2368 \\
			STF & & \ding{51} & 31.601 & 38.137 & \underline{68.694} & 20.153 & 663 \\
			\rowcolor{tblue} \textbf{UKF} & & \ding{51} & \underline{33.201} & \underline{41.644} & 68.451 & \underline{22.261} & \underline{609} \\
			\rowcolor{tblue} \textbf{UKF$^\dagger$}  & & \ding{51} & \textbf{34.067} & \textbf{44.643} & \textbf{68.958} & \textbf{23.594} & \textbf{544} \\
			\bottomrule
		\end{tabular}%
	}
\end{table}

\section{Conclusion}
In this paper, we introduces a unified underwater MFT benchmark and a specialized tracking framework for fish morphology and erratic swimming patterns. Our baseline achieves state-of-the-art performance with 34.1 HOTA, significantly outperforming other trackers. Statistical analysis reveals fundamental differences between fish and terrestrial tracking scenarios, highlighting the necessity for specialized underwater approaches. However, significant challenges remain in handling visually similar fish appearances, extreme density scenarios, and highly erratic swimming patterns.

\section*{Appendix.A Appearance Feature Analysis}
\label{sec:feature_analysis}

To further quantify the challenges posed by underwater fish tracking, we analyze the distribution and separability of identity-specific appearance features extracted from MFT25 in comparison to the widely used pedestrian dataset MOT17. We extract per-track feature embeddings using a pretrained ReID model and visualize their distributions using t-SNE projections and cosine similarity heatmaps.

\begin{figure}[htbp]
	\centering
	\includegraphics[width=\linewidth]{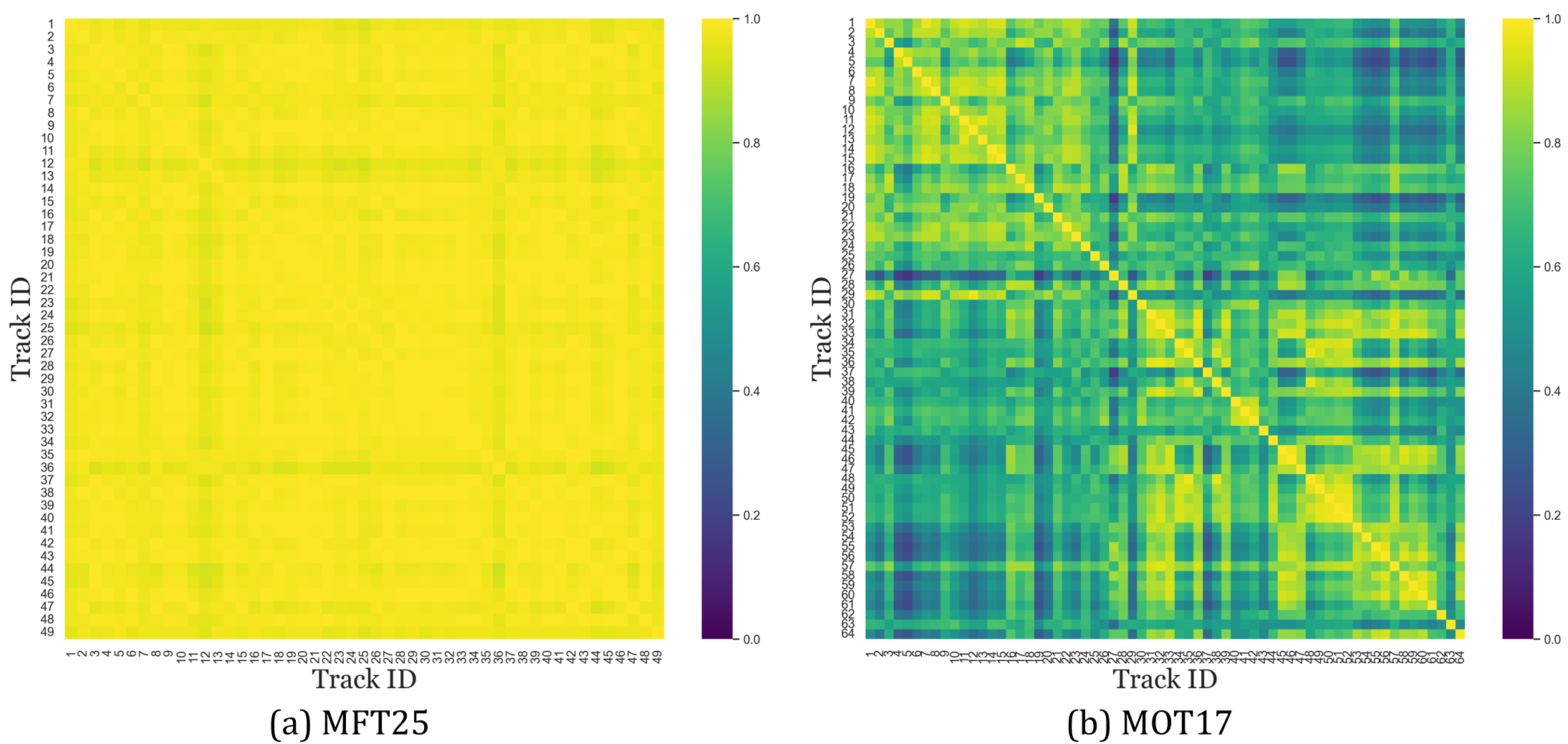}
	\caption{Pairwise cosine similarity between track-level appearance features. MFT25 shows uniformly high inter-track similarity, while MOT17 exhibits clearer identity separation.}
	\label{fig:heatmap}
\end{figure}

A complementary analysis is shown in Fig.~\ref{fig:heatmap}, which depicts pairwise cosine similarities across all track embeddings. In MFT25, the heatmap reveals uniformly high similarity scores across identities, indicating limited discriminative cues for visual association. In contrast, MOT17 exhibits a more diagonal-dominant structure, where self-similarity (along the diagonal) is clearly distinguishable from inter-class similarity, reflecting stronger feature separability. These observations underscore the difficulty of re-identification in underwater settings and motivate the need for enhanced appearance modeling techniques specifically tailored to low-contrast, deformable targets.

\begin{figure}[htbp]
	\centering
	\includegraphics[width=1\linewidth]{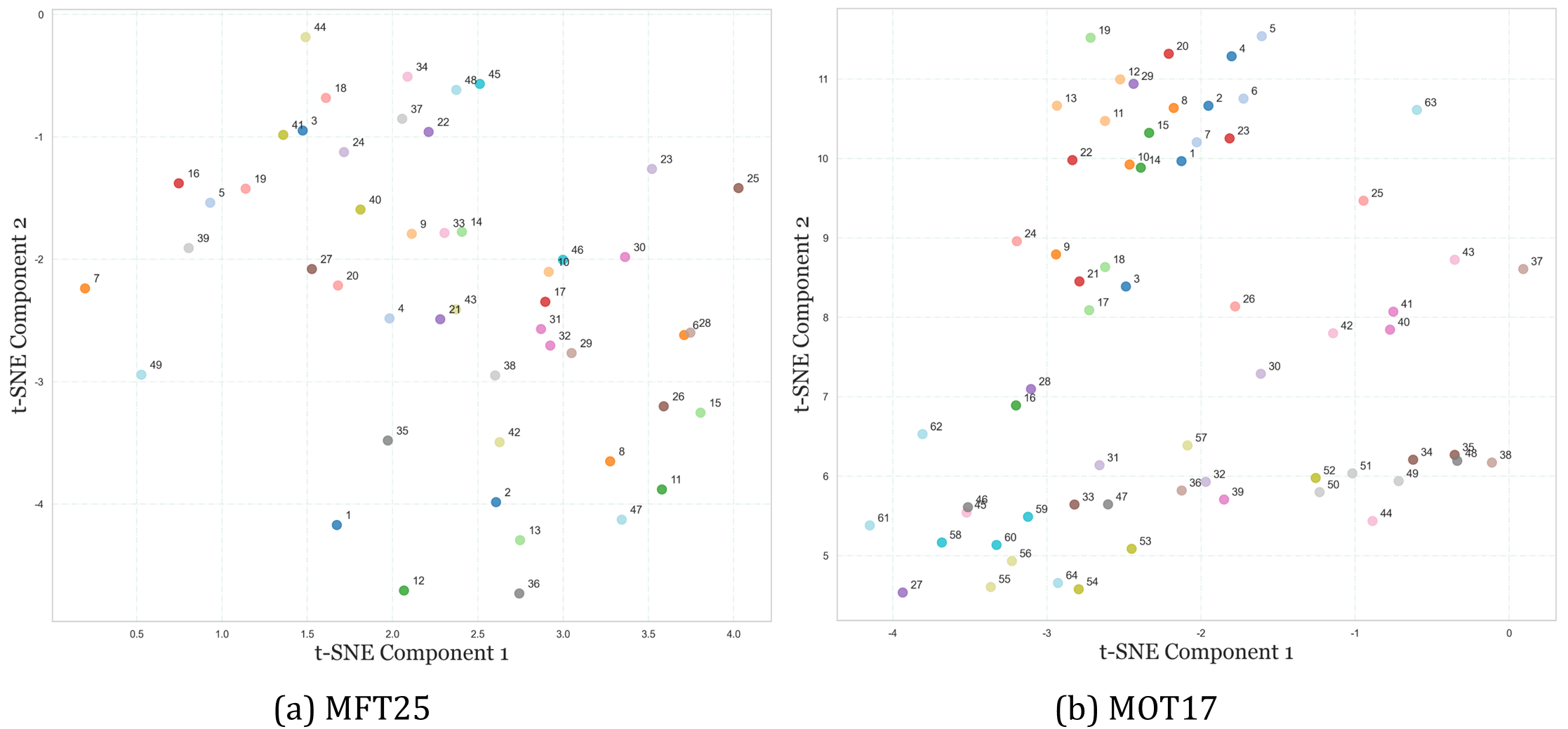}
	\caption{t-SNE projection of track-level appearance embeddings. Identity clusters in MFT25 are less separable compared to those in MOT17, reflecting greater visual ambiguity.}
	\label{fig:tsne}
\end{figure}

As illustrated in Fig.~\ref{fig:tsne}, the track-level embeddings in MFT25 (left) exhibit significantly higher intra-class overlap and reduced inter-class margins compared to MOT17 (right), where distinct identity clusters are more well-formed and linearly separable. This suggests that fish exhibit substantially less distinctive visual signatures, likely due to their similar body shapes, colorations, and the frequent self-occlusions arising from dense schooling behaviors.

\section*{Appendix.B Statistical Analysis}
\label{sec:sa}

Building upon the comparative statistical analysis of motion direction across diverse tracking datasets presented in the main paper, we conducted additional experiments to quantify the distinctive motion characteristics of fish tracking. Our analysis included pedestrians~\cite{dendorfer2021motchallenge}, dancers~\cite{sun2022dancetrack}, bees~\cite{cao2025topic}, and cells~\cite{anjum2020ctmc}. These experiments aim to differentiate fish tracking from other domains based on their unique motion patterns.

\begin{figure*}[!t]
	\centering
	\includegraphics[width=1\linewidth]{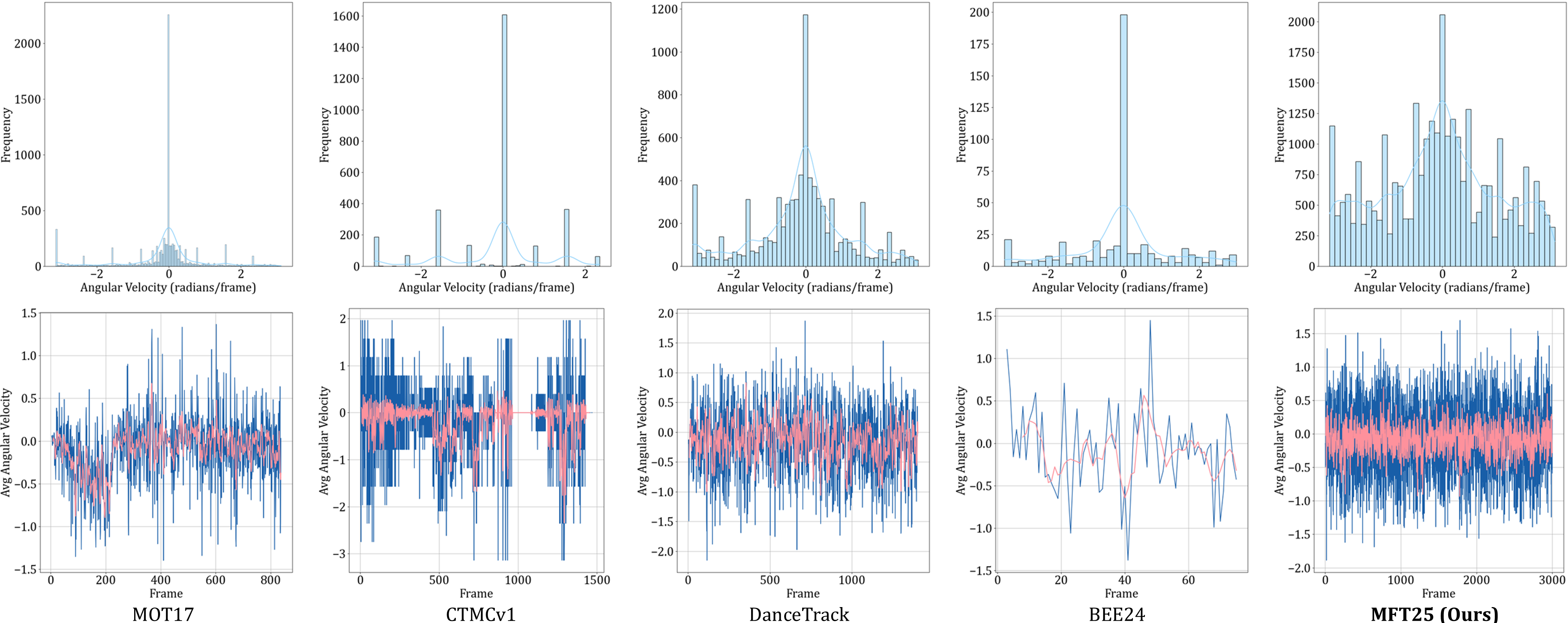}
	\caption{Temporal evolution of angular velocities across different tracking datasets. MFT25 demonstrates significantly higher directional variability and more erratic motion patterns compared to other tracking scenarios.}
	\label{fig:angular-speed}
\end{figure*}

Fig.~\ref{fig:angular-speed} illustrates the angular velocity distribution across different datasets. Our analysis revealed statistically significant differences in motion patterns between domains. The MOT17~\cite{dendorfer2021motchallenge} dataset demonstrates highly stable angular velocities with predominantly linear motion patterns characteristic of pedestrian movement. Conversely, the MFT25 dataset exhibits substantially greater angular velocity variations and more erratic directional changes while maintaining high swimming speeds. This unpredictable behavior results from the hydrodynamic properties of underwater motion and environmental factors, which generate inherently nonlinear motion trajectories.

\begin{figure*}[!t]
	\centering
	\includegraphics[width=1\linewidth]{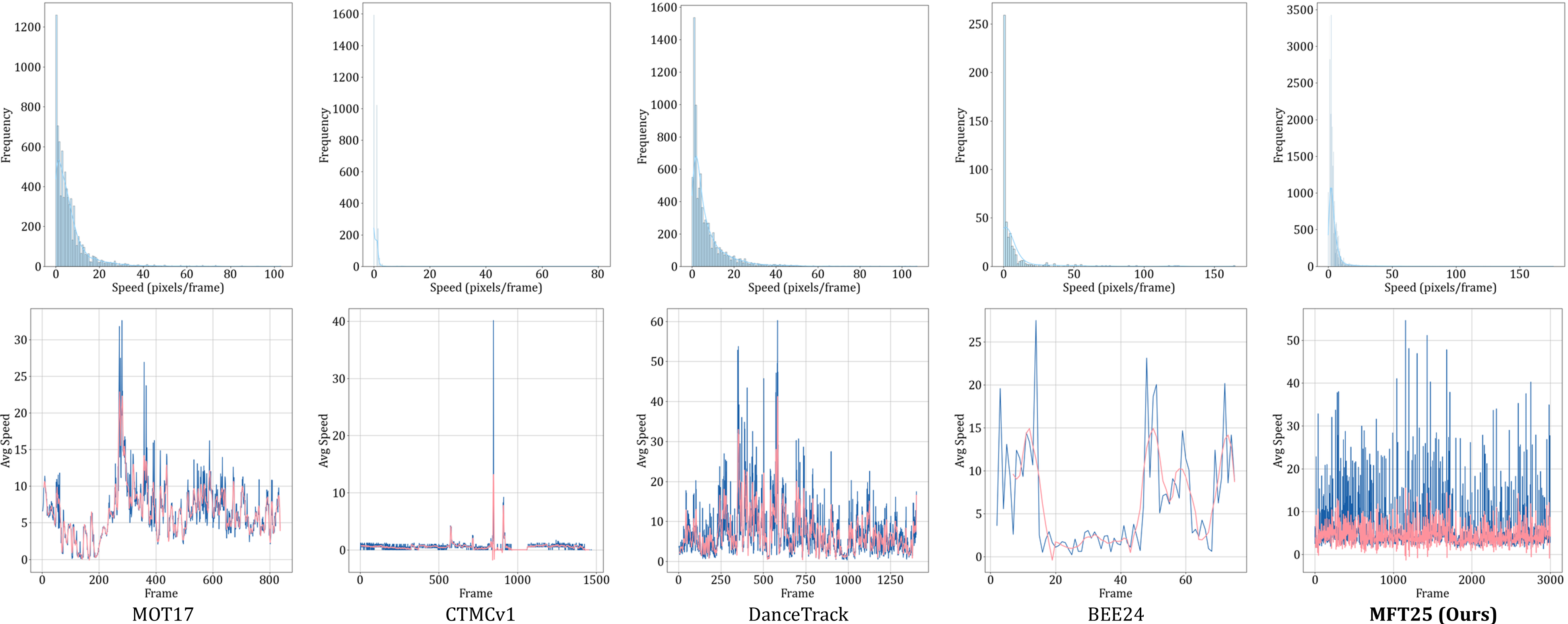}
	\caption{Temporal evolution of average object speeds across different tracking datasets. MFT25 demonstrates consistently higher velocities and greater variability compared to other tracking scenarios.}
	\label{fig:speed-analysis}
\end{figure*}

As illustrated in Fig.~\ref{fig:speed-analysis}, the MFT25 dataset demonstrates substantial fluctuations in average motion speed, whereas the CTMC~\cite{anjum2020ctmc} dataset exhibits relatively stable speed profiles. These fundamental kinematic differences underscore why general-purpose tracking algorithms frequently underperform in underwater environments and emphasize the critical need for specialized fish tracking methodologies capable of adapting to these distinctive motion characteristics.

\section*{Appendix.C Per-Sequence Performance}
\label{sec:clips-analysis}

Table~\ref{tab:sequence-performance} presents a comprehensive evaluation of our SU-T baseline across individual sequences in the MFT25 dataset, revealing substantial heterogeneity in tracking difficulty and performance characteristics. The results demonstrate significant variation across sequences, with HOTA scores ranging from 20.3 to 65.2, highlighting the diverse challenges inherent in underwater fish tracking scenarios.

\begin{table*}[!t]
	\caption{Performance of our baseline method on individual sequences in the MFT25 dataset.}
	\label{tab:sequence-performance}
	\centering
	\resizebox{\linewidth}{!}{%
		\begin{tabular}{lccccccccccccccc}
			\toprule
			\textbf{Clip}      & \textbf{HOTA↑} & \textbf{IDF1↑} & \textbf{MOTA↑} & \textbf{MOTP↑} & \textbf{DetA↑} & \textbf{DetPr↑} & \textbf{DetRe↑} & \textbf{AssA↑} & \textbf{AssPr↑} & \textbf{AssRe↑} & \textbf{IDs↓} & \textbf{IDFP↓} & \textbf{IDFN↓} & \textbf{LocA↑} & \textbf{Frag↓} \\ \midrule
			BT-002         &     35.961     &     58.562     &     55.923     &     60.727     &     45.139     &     53.968      &     52.020      &     28.788     &     43.979      &     35.754      &      35       &      1917      &      2095      &     68.930      &      128       \\
			BT-004         &     51.244     &     68.933     &     93.028     &     69.256     &     62.416     &     69.058      &     68.071      &     42.125     &     59.115      &     47.578      &      21       &      2299      &      2408      &     75.671     &       59       \\
			MSK-003         &     34.188     &     53.067     &     50.393     &     60.078     &     41.026     &     52.216      &     47.920      &     28.930     &     44.874      &     35.650      &      134      &     12488      &     15001      &     68.854     &      646       \\
			PF-002         &     65.162     &     98.650     &     97.300     &     69.890     &     64.927     &     70.203      &     70.226      &     65.399     &     70.729      &     70.400      &       0       &       82       &       80       &     76.067     &       18       \\
			SN-009         &     27.169     &     28.799     &     82.766     &     70.719     &     58.500     &     67.828      &     65.441      &     12.692     &     44.586      &     15.470      &      173      &     29339      &     30853      &     76.148     &      431       \\
			SN-011         &     20.264     &     26.165     &     47.423     &     64.313     &     40.045     &     48.522      &     53.843      &     10.421     &     41.176      &     13.032      &      161      &     27146      &     23575      &     71.909     &      522       \\
			SN-014         &     40.046     &     47.718     &     84.339     &     72.997     &     60.300     &     70.370      &     67.023      &     26.700     &     61.338      &     29.128      &      83       &      9840      &     10802      &     77.814     &      202       \\ \midrule
			\rowcolor{tblue} Total &     33.351     &     41.717     &     68.450     &     67.350     &     49.943     &     59.947      &     59.243      &     22.425     &     49.482      &     26.206      &      607      &     83111      &     84814      &     73.911     &      2006      \\ \bottomrule
		\end{tabular}%
	}
\end{table*}

\paragraph{Optimal Conditions.}
The PF-002 sequence emerges as an exemplary case, achieving exceptional performance with 65.2 HOTA, 98.7 IDF1, and remarkably zero identity switches. This outstanding performance can be attributed to several favorable environmental and behavioral factors. The sequence features distinct fish appearances with high inter-individual visual variation, enabling robust feature discrimination. Additionally, the predictable movement patterns and minimal occlusions create ideal conditions for consistent trajectory estimation. The near-perfect IDF1 score of 98.7, combined with the absence of identity switches, demonstrates that under optimal conditions, the SU-T approach can achieve human-level tracking accuracy. The low fragmentation count (18) and minimal false positives (82) and false negatives (80) further corroborate the robustness of the tracking framework in favorable underwater environments.
The BT-004 sequence represents another success case, achieving strong results with 51.2 HOTA and 68.9 IDF1. This sequence demonstrates robust tracking capabilities in moderately challenging scenarios where fish maintain relatively consistent swimming behaviors. The balanced DetA (62.4) and AssA (42.1) scores indicate that both detection and association components perform adequately, suggesting that the sequence contains manageable levels of visual complexity and motion predictability. The moderate number of identity switches (21) and reasonable fragmentation (59) demonstrate the method's resilience to typical underwater tracking challenges.

\paragraph{Environmental Complexity.}
The MSK-003 sequence exhibits moderate performance with 34.2 HOTA but reveals specific challenges in identity management, suffering from a high number of identity switches (134 IDs). This pattern suggests the presence of crowded underwater environments where multiple fish exhibit similar appearances or engage in frequent close interactions leading to identity confusion. The substantial IDFP (12,488) and IDFN (15,001) values indicate significant challenges in maintaining consistent detections, likely due to dynamic lighting conditions or dense fish populations. The high fragmentation count (646) suggests frequent track interruptions, possibly caused by temporary occlusions or fish moving in and out of the camera's field of view.

\paragraph{Challenging Conditions.}
The SN-009 and SN-011 sequences represent the most challenging scenarios in the dataset, with HOTA scores of 27.2 and 20.3, respectively. These sequences demonstrate substantial identification failures, with exceptionally high IDFN values of 30,853 and 23,575, revealing fundamental limitations of current tracking approaches in visibility-constrained underwater scenarios.
SN-011, in particular, shows the most degraded performance with 20.3 HOTA, attributed to challenging conditions including frequent occlusions and highly erratic fish movements. The extremely low AssA score (10.4) indicates severe difficulties in data association, suggesting that traditional appearance-based and motion-based association strategies fail under these conditions. The high number of false negatives (23,575) implies significant detection failures, potentially due to poor water clarity, extreme lighting variations, or camouflaging effects where fish blend with the background environment.
SN-009 exhibits similar challenges with a notably low IDF1 score of 28.8, indicating poor identity preservation throughout the sequence. The substantial difference between DetA (58.5) and AssA (12.7) suggests that while detection performance remains reasonable, the association component fails dramatically, likely due to rapid appearance changes or highly unpredictable movement patterns that violate motion model assumptions.

\paragraph{Detection vs. Association.}
A critical insight emerges from examining the relationship between detection accuracy (DetA) and association accuracy (AssA) across sequences. High-performing sequences (PF-002, BT-004) maintain balanced DetA and AssA scores, indicating that both components contribute effectively to overall performance. In contrast, challenging sequences (SN-009, SN-011) show substantial disparities, with DetA significantly outperforming AssA, suggesting that while individual detections remain feasible, temporal coherence and identity maintenance become the primary bottlenecks.
The precision-recall balance also varies significantly across sequences. Sequences like PF-002 achieve excellent balance with DetPr (70.2) closely matching DetRe (70.2), while problematic sequences show imbalanced patterns. For instance, SN-011 exhibits higher recall (53.8) than precision (48.5), indicating a tendency toward over-detection that subsequently complicates the association process.

\paragraph{Fragmentation and Temporal Consistency.}
The fragmentation analysis reveals important insights about temporal tracking consistency. PF-002's minimal fragmentation (18) contrasts sharply with MSK-003's extensive fragmentation (646), highlighting how environmental complexity directly impacts trajectory continuity. High fragmentation correlates strongly with challenging conditions, suggesting that improving track continuation strategies could significantly enhance overall performance in difficult scenarios.

\begin{figure*}[tbp]
	\centering
	\includegraphics[width=\linewidth]{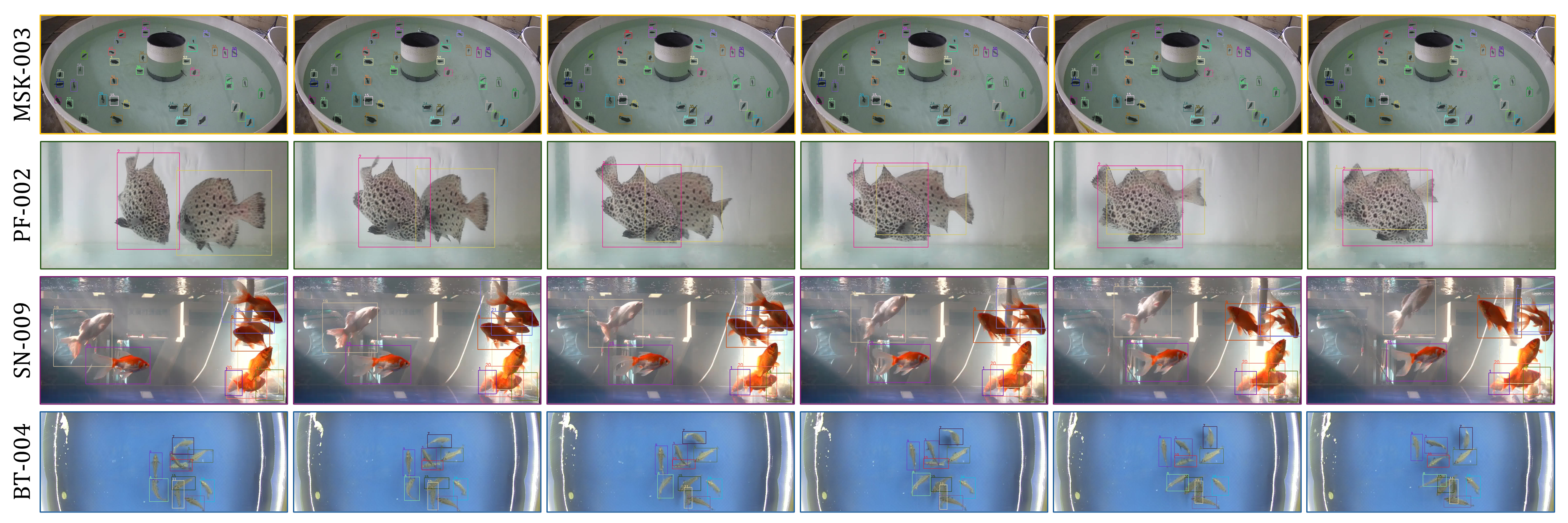}
	\caption{Visualization of SU-T tracking performance across diverse scenarios in the MFT25 dataset. Each tracked fish maintains consistent identity despite challenging underwater conditions including variable visibility, population density, and complex backgrounds. Best viewed in color.}
	\label{fig:vis-ours}
\end{figure*}

Finally, to visualize the practical effectiveness of SU-T, Fig.~\ref{fig:vis-ours} presents tracking results across diverse scenarios in the MFT25 dataset. The visualizations demonstrate that SU-T maintains consistent identity assignments despite varying challenges including different fish densities, species morphologies, and environmental conditions.

\section*{Appendix.D Hyperparameters}
\label{sec:hyperparams}

We know that the tuning of hyperparameters has a significant impact on experimental results. To ensure robust detection and tracking in challenging underwater fish tracking conditions, we configured task-specific hyperparameters for both the training and inference pipelines. Notably, these hyperparameters were tuned to achieve the best performance on the MFT25 dataset and were kept consistent across all experiments in the main paper to ensure a fair and unbiased comparison.

\paragraph{Detection Configuration.} The detection model employs a depth multiplier of 1.33 and width multiplier of 1.25 to enhance representation capacity for fine-grained fish features. We utilize an input resolution of $800 \times 1440$, which aligns with the wide-format aspect ratio commonly observed in underwater recordings. Multi-scale training is applied with random image sizes ranging from 18 to 32 grid units. The model is trained for 80 epochs using SGD with an initial learning rate of $0.001/64$, scheduled with 1 warm-up epoch and 10 final epochs without augmentation to stabilize training. Each image is normalized using ImageNet statistics, with mosaic and mixup augmentations enabled except during the no-augmentation phase.

\paragraph{Tracking Configuration.} We employ a two-stage tracking scheme that combines IoU-based association with ReID embedding similarity. The primary association utilizes a modified IoU metric tailored for fish morphology, denoted as FishIoU, with a threshold of 0.25. A ByteTrack-style secondary association is enabled, with both stages integrated into a temporal consistency module (TCM) using equal weights of 1.0. Track motion smoothing is achieved using an inertia coefficient of 0.05 to accommodate abrupt directional changes while preserving trajectory stability.
To enhance identity preservation, we incorporate a ResNet-101-IBN backbone trained specifically on MFT25. ReID-based long-term correction is triggered when similarity exceeds a threshold of 0.4, with confidence-aware embedding gains set to 1.3 for high-confidence detections and 1.2 otherwise.

\paragraph{Evaluation Protocol.} Evaluation is performed every 5 epochs using COCO-style AP metrics. Detection confidence is thresholded at 0.1, and non-maximum suppression (NMS) is applied with an IoU threshold of 0.7.

\section*{Acknowledgments}
The paper is supported in part by Beijing Smart Agriculture Innovation Consortium Project (BAIC10-2025). The authors gratefully acknowledge National Innovation Center for Digital Fishery  - China Agricultural University, Key Laboratory of Agricultural Informatization Standardization - MARA, P. R. China, Key Laboratory of Smart Farming Technologies for Aquatic Animals and Livestock - MARA, P. R. China, National Innovation Center for Digital Agricultural Products Circulation - MARA, P. R. China, and State Key Laboratory of Efficient Utilization of Agricultural Water Resources - China Agricultural University. Weiran Li gratefully acknowledges financial support from the China Scholarship Council (No. 202406350102).

{
	\small
	\bibliographystyle{ieeenat_fullname}
	\bibliography{main}
}

\newpage

\end{document}